\documentclass[10pt,twocolumn,letterpaper]{article}

\usepackage{iccv}
\usepackage{times}
\usepackage{epsfig}
\usepackage{graphicx}
\usepackage{amsmath}
\usepackage{amssymb}
\usepackage{multirow}
\usepackage{tabulary}
\usepackage{colortbl}
\usepackage{booktabs}
\usepackage[breaklinks=true,bookmarks=false]{hyperref}

\iccvfinalcopy % *** Uncomment this line for the final submission

\def\iccvPaperID{****} % *** Enter the ICCV Paper ID here

\ificcvfinal\fi

\begin{document}

%%%%%%%%% TITLE
\title{Learning Meta-class Memory for Few-Shot Semantic Segmentation}

\author{Zhonghua Wu$^{1,2}$~~~~Xiangxi Shi$^{3}$~~~~Guosheng Lin\thanks{Corresponding author: G. Lin (e-mail: {\tt gslin@ntu.edu.sg})}~~$^{1,2}$~~~~Jianfei Cai$^{4}$\\
 $^{1}$S-lab, Nanyang Technological University \\$^{2}$School of Computer Science and Engineering, Nanyang Technological University\\$^{3}$Electrical Engineering and Computer Science, Oregon State University\\$^{4}$Dept of Data Science and AI, Monash University \\
{\tt\small zhonghua001@e.ntu.edu.sg~~ shixia@oregonstate.edu~~gslin@ntu.edu.sg~~jianfei.cai@monash.edu}
}

\maketitle
\ificcvfinal\thispagestyle{empty}\fi

%%%%%%%%% ABSTRACT
\begin{abstract}
    Currently, the state-of-the-art methods treat few-shot semantic segmentation task as a conditional foreground-background segmentation problem, assuming each class is independent. In this paper, we introduce the concept of meta-class, which is the meta information (e.g. certain middle-level features) shareable among all classes. To explicitly learn meta-class representations in few-shot segmentation task, we propose a novel Meta-class Memory based few-shot segmentation method (MM-Net), where we introduce a set of learnable memory embeddings to memorize the meta-class information during the base class training and transfer to novel classes during the inference stage. Moreover, for the $k$-shot scenario, we propose a novel image quality measurement module to select images from the set of support images. A high-quality class prototype could be obtained with the weighted sum of support image features based on the quality measure. Experiments on both PASCAL-$5^i$ and COCO dataset shows that our proposed method is able to achieve state-of-the-art results in both 1-shot and 5-shot settings. Particularly, our proposed MM-Net    achieves 37.5\% mIoU on the COCO dataset in 1-shot setting, which is 5.1\% higher than the previous state-of-the-art.
\end{abstract}

%%%%%%%%% BODY TEXT

\begin{figure}[t]
\centering
    \includegraphics[width=\linewidth]{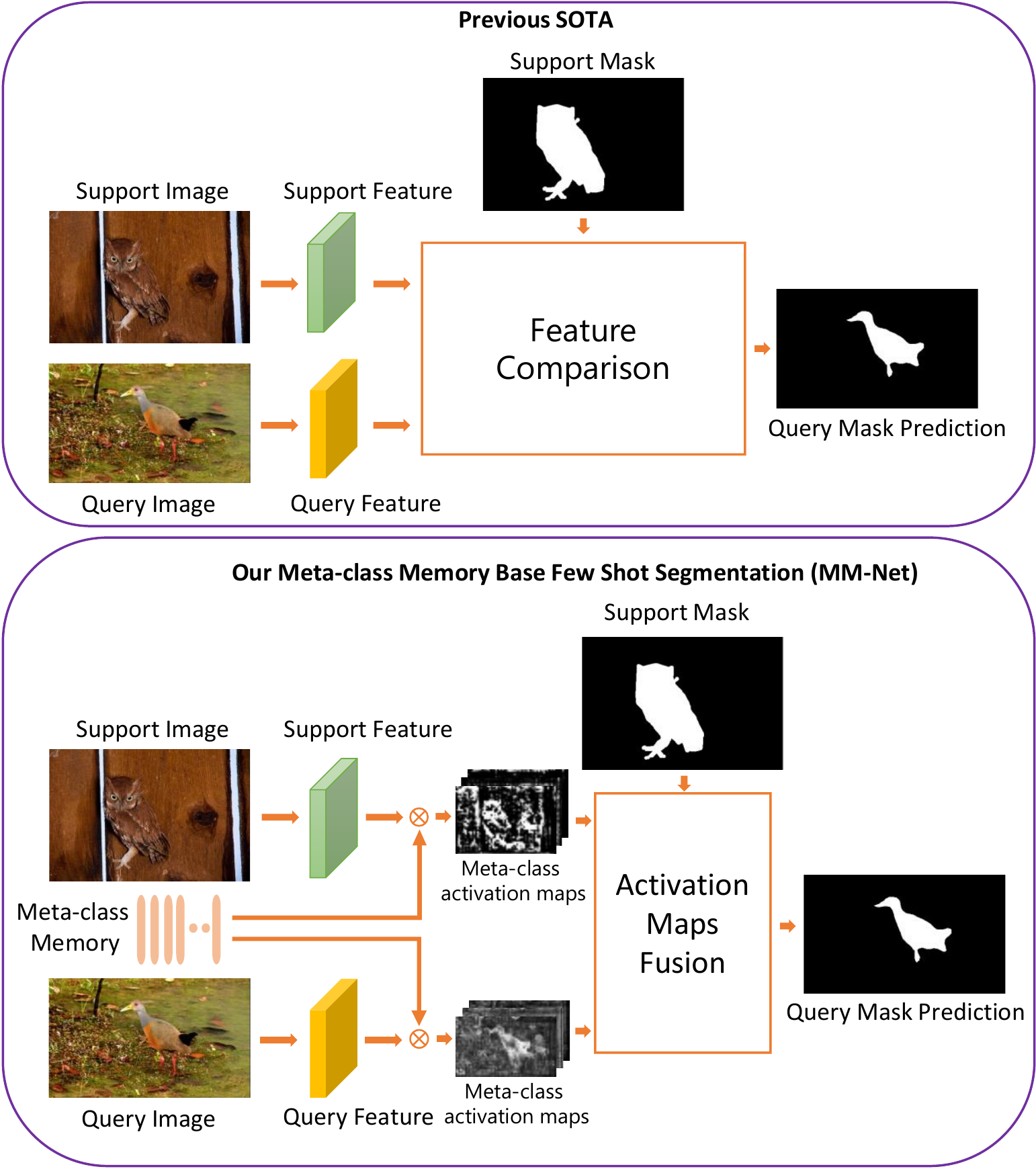}
    \caption{Comparison between the typical pipeline of state-of-the-art (SOTA) methods (top part) and that of our proposed meta-class memory based network (MM-Net) (bottom part) for few-shot segmentation. The main difference is that SOTA treats the task as a class-agnostic conditional foreground-background segmentation problem, while we propose to learn a set of meta-class middle-level representations  shareable between base and novel classes.}
    \label{first_diagram}
\end{figure}

\section{Introduction}
With the development of convolution neural networks (CNNs), fully supervised image semantic segmentation \cite{lin2017refinenet, chen2017rethinking} has achieved great success in both speed and accuracy. However, the state-of-the-art image segmentation methods usually require abundant pixel-level annotations which requires huge human labeling efforts. If we want to segment a new class that has not been seen in the training set, we usually need to label thousands of images for the new class. In order to reduce human labeling efforts on novel classes, the few-shot image segmentation task \cite{zhang2019canet, liu2020crnet} has been introduced, which aims to predict the segmentation mask of a query image of a novel class 
with only one or a few labelled support images in testing, while with abundant images of base classes with full annotations in training.

The top part of Fig.~\ref{first_diagram} shows the typical pipeline of the state-of-the-art (SOTA) few-shot image segmentation methods \cite{zhang2019canet, zhang2019pyramid}.
Firstly, a pre-trained CNN network is used to extract the features of both support and query images. Then, the two features are typically processed by convolutional layers and compared for similarity so as to generate the segmentation map for the query image. Essentially, these methods treat the few-shot segmentation task as a conditional binary foreground-background segmentation problem, i.e. to find and segment the most relevant regions in the query image based on the given support images and their masks, regardless of the class information. 

The class-agnostic design in SOTA is understandable. This is because by class-agnostic design the interaction / comparison between query and support features in base classes can be transferred to novel classes. However, we argue that although  different classes of objects are quite different, there are still some common attributes or  middle-level knowledge shareable among them, which we call meta-class information. Similar observations have been made in \cite{zou2020revisiting, huang2019all, wu2020exploring} for classification and detection tasks, where some low-level information (e.g. circle, dot) and middle-level information (e.g. wings, limbs) are shared among different classes.

Motivated by this, in this paper, we propose a novel Meta-class Memory Module (MMM) to learn middle-level meta-class embeddings shareable between base and novel classes for few-shot segmentation. As shown in the lower part of Fig.~\ref{first_diagram}, a set of meta-class memory embeddings is introduced into the SOTA pipeline, which can be learned through the back-propagation during the base class training. The meta-class memory embeddings are then used to attend the middle-level features of query and support images to obtain meta-class activation maps. This can be considered as aligning both query and support middle-level features to the meta-class embeddings. Based on the obtained meta-class activation maps of query and support images, we then perform interaction / comparison between them to propagate the support mask information from support activation maps to query activation maps, and the fused query activation maps are finally used for the query mask prediction.

In addition, when it comes to the $k$-shot scenarios, which means more than one support images are given, previous methods usually apply an average operation \cite{zhang2019pyramid} on the few support image features to obtain the  class prototype feature. However, we observe that some support images are in low quality which is hard to represent the support class. Thus, we further propose a Quality Measurement Module (QMM) to obtain the quality measure for each support image. Based on the quality measure, the features from all support images are fused via weighted sum to get a better class prototype feature.

In our experiments, we follow the design of~\cite{tian2020pfenet} to perform training and testing for a fair comparison. We evaluate our proposed method on four different splits with 1-shot and 5-shot settings on PASCAL-$5^i$~\cite{shaban2017one} and COCO~\cite{lin2014microsoft} datasets. 
Our method is able to achieve state-of-the-art results on both datasets under both 1-shot and 5-shot settings. 

Our main contributions can be summarized as follows:
\begin{itemize}
\item For few-shot semantic image segmentation, to our knowledge, we are the first one to introduce a set of learnable embeddings to memorize the meta-class information during base class training that can be transferred to novel classes during testing.
Specifically, a Meta-class Memory Module (MMM) is proposed to generate the meta-class activation maps for both support and query images, which is helpful for the final query mask prediction. 

\item For $k$-shot scenarios, a Quality Measurement Module (QMM) is proposed to measure the quality of all the support images so as to effectively fuse all the support features. 
With QMM, our model is able to pay more attention to the high quality support samples for better query image segmentation.

\item Extensive experiments on PASCAL-$5^i$ and COCO datasets show that our proposed method performs the best in all settings. 
Specifically, our method significantly outperforms SOTA 
on the large scale dataset COCO, with 5.1\% mean mIoU gain,   
as our memory embeddings are able to learn a universal meta-class representation. 

\end{itemize}

\begin{figure*}[t]
\centering
    \includegraphics[width=0.8\linewidth]{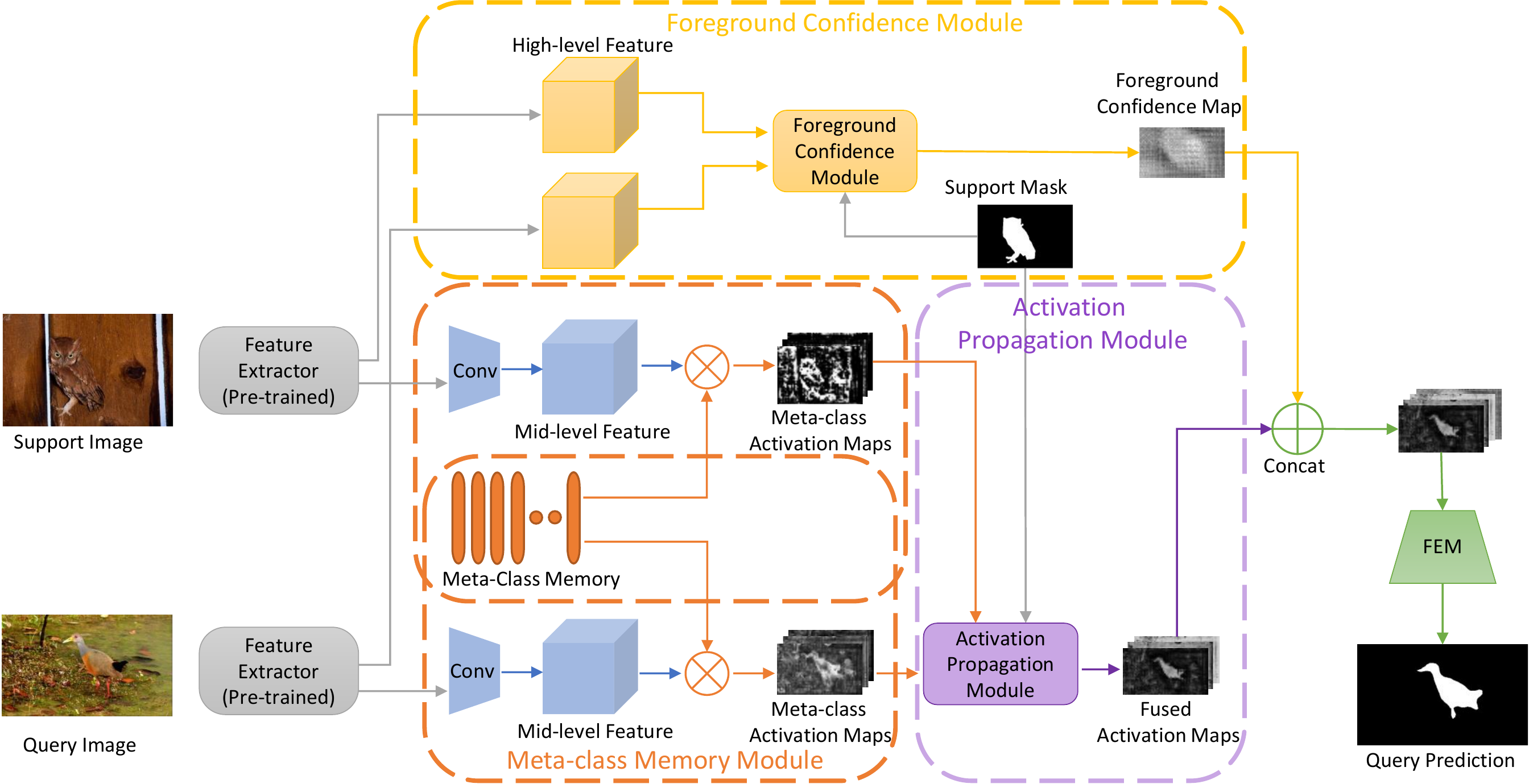}
    \caption{Overview of our proposed meta-class memory based network (MM-Net) for few-shot semantic segmentation. Different from previous few-shot segmentation methods, Meta-class Memory Module (MMM) (orange) is introduced to learn the meta-class features that can be shared among all base and novel classes and generate meta-class activation maps  for support and query images, respectively. Then,  Activation Propagation Module (APM) (purple) is used to propagate support mask information to the query activation maps for the query mask generation. 
    Meanwhile, foreground confidence module (FCM) (yellow) is used to obtain a confidence map from the high-level image features. Finally, the fused query activation maps are concatenated with the foreground confidence map and fed into FEM \cite{tian2020pfenet} for the final query segmentation mask prediction (green).}
    \label{overall}
\end{figure*}

\section{Related Works}
\subsection{Semantic Segmentation} 
Semantic segmentation~\cite{wu2019keypoint, zhang2020splitting, liu2020weakly, liu2020guided} is a task of classifying each pixel in an image into a specified category and has been applied in various fields~\cite{wu2019m2e, shi2021remember}. State-of-the-art segmentation methods are usually based on the Fully Convolutional Network (FCN)~\cite{long2015fully}, which uses a classification network as the backbone and replaces fully connected layers with convolutional layers to predict the dense segmentation map. Later, to obtain a higher resolution prediction and have a larger receptive field of the network, DeepLab \cite{chen2014semantic, chen2017deeplab} proposed to use dilated convolutions which insert holes to the convolutional filters instead of using the conventional convolution with downsampling. Recently, Chen et al. further explored the effect of atrous convolutions, multi-grid, atrous spatial pyramid pooling, difference backbones, and different training sizes in DeepLab V3 \cite{chen2017rethinking} and DeepLab V3+ \cite{chen2018encoder}. These methods usually require abundant pixel-level annotations for all the classes during training and cannot generalize to novel classes with only a few labelled images.

\subsection{Few-Shot Semantic Segmentation}
Few-shot semantic segmentation \cite{fan2020fgn, gairola2020simpropnet, wang2020few, ouyang2020self, shaban2017one, dong2018few} aims to give a dense segmentation prediction for new class query images with only a few labeled support images. CANet \cite{zhang2019canet} proposed Dense Comparison Module (DCM) and Iterative Optimization Module (IOM) to give a dense prediction and iteratively refine the prediction. Similarly, the prototype alignment regularization was used in PANet~\cite{wang2019panet} which encourages the model to learn more consistent embedding prototypes. Later, PGNet \cite{zhang2019pyramid} used a Graph Attention Unit (GAU) to build the local similarity between support and query images. Liu et al.~\cite{liu2020crnet} proposed to use a Siamese Network on query and support images to get the co-occurred features between the two images. More recently, following the practice in PGNet~\cite{zhang2019pyramid} which uses the pyramid structure to refine results, PFENet~\cite{tian2020pfenet} used a multi-scale decoder Feature Enrichment Module (FEM) to incorporate the prior masks and query features to give a better segmentation map prediction. 

Unlike all the existing few-shot semantic segmentation methods, we introduce a concept of meta-class memory that could learn a set of shareable meta-class representations among base and novel classes. 

\subsection{Meta Learning}
Most state-of-the-art recognition methods
require a large number of training images with abundant annotations which often need tremendous human labeling efforts. Meta-learning methods \cite{finn2017model}, also known as learning to learn, have been introduced to better transfer existing knowledge to novel classes or get faster training on new given data. One popular set of  approaches~\cite{li2017meta, rusu2018meta} are to learn a meta-learner which could help deep neural networks optimize faster when given new data on unseen classes. Another set of meta-learning approaches~\cite{finn2017model} are introduced to learn a better parameter initialization which could be fast to optimize with fewer training data. Metric learning methods~\cite{vinyals2016matching} are belonging to another type that use a certain similarity measurement to obtain the classification results over different classes. On the other hand, Munkhdalai et al.~\cite{munkhdalai2017meta} proposed an external memory model across different tasks and can shift its inductive biases via fast parameterization for rapid generalization on new tasks.

Different from these meta-learning approaches, we do not have a meta-learner. Instead, we construct a meta-class memory to capture representative middle-level features for better transfer between base and novel classes for few-shot semantic segmentation.

\section{Method}
Fig.~\ref{overall} gives an overview of our proposed meta-class memory based network (MM-Net) for few shot semantic segmentation. It consists of three major modules, meta-class memory module (MMM), activation propagation module (APM) and foreground confidence module (FCM), as well as two off-the-shelf modules, feature extraction backbone and feature enrichment module (FEM) \cite{tian2020pfenet}. MMM is particularly novel, which learns the meta-class features that can be shared among all base and novel classes and generate meta-class activation maps for support and query images, respectively. APM is used to propagate support mask information to the query activation maps for the query mask generation. FCM is to retain the conventional interactions of the high-level features of the query and support images. Moreover, we also propose an additional Quality Measurement Module (QMM) to measure the quality of different support images in $k$-shot settings so as to fuse the information from different support images in a better way. In the following, we describe the major modules of our MM-Net in detail.

\subsection{Meta-class Memory Module}
Meta-class Memory module (MMM) aims to learn the meta-class information that can be shared among all classes and use them to encode / classify the middle-level features of the query or support image. As shown in Fig.~\ref{overall}, the input to MMM includes the features of either query image $I_Q$ or support image $I_S$, and a set of meta-class embeddings {$M^1 ... M^N$}, each with a dimension of $D$, where we set $N$ as 50 and $D$ as 256. These meta-class embeddings can be learned during the network training through the back propagation. The outputs of MMM are the meta-class activation maps corresponding to the given image. 
In particular, we firstly use the ResNet50 as a feature extractor to extract the features for both support and query images. Similar to the previous few-shot segmentation works~\cite{zhang2019canet,zhang2019pyramid}, the feature extractor is pre-trained on the image classification task. We choose the features from the 2nd and 3rd levels, since the middle-level features are better for transfer, same as the observation in~\cite{zhang2019canet}. Then, we apply a channel-wise concatenation of the 2nd and 3rd level features, followed by a $3 \times 3$ convolution layer to get the feature maps $F_Q$ and $F_S$ for the query and support images, respectively. Then, we compute the similarity between the meta-class memory $M$ and image feature maps $F$ to get the meta-class activation maps $Act_Q$ and $Act_S$ for query and support images,  respectively:
\begin{equation}
Act^{n}(x,y) = \sigma(F(x, y)^TM^{n}), 
\label{eq:act}
\end{equation}
where $Act^{n}(x,y)$ indicates the $n$th meta-class activation map at the spatial location (x, y) obtained by the $n$th embedding, and $\sigma(\cdot)$ is the Sigmoid function to normalize the value between 0 to 1.

\begin{figure}[t]
\centering
    \includegraphics[width=0.65\linewidth]{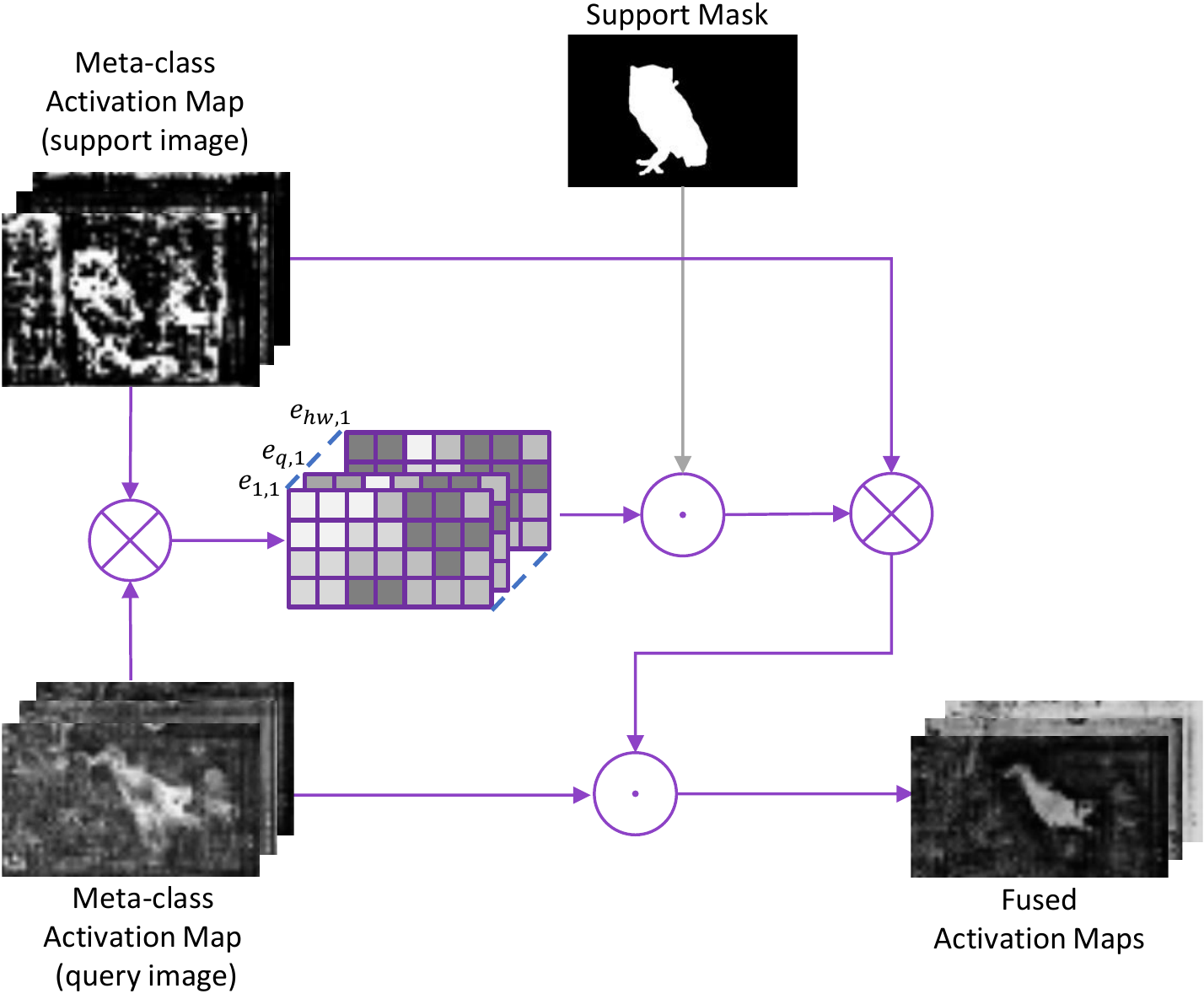}
    \caption{Illustration of Activation Propagation Module (APM).}
    \label{APM}
\end{figure}

\subsection{Activation Propagation Module} \label{Sec:APM}
With the two memory activation maps $Act_S$ and $Act_Q$ encoded by the $N$ meta-class embeddings, which are of the dimensions of $H\times W \times N$ and $H\times W$ is the spatial dimensions, the purpose of this activation propagation module (APM) is to propagate the label information (support mask) from the support image to the unlabeled query image for the mask generation. 

For APM, we adopt an approach similar to~\cite{zhang2019pyramid} but using our unique memory activation maps, where we treat the vector at each spatial location of the activation maps as a node. Fig.~\ref{APM} illustrates the process of APM. 
In particular, we denote $h_q \in Act_Q$ as a query node and $h_s \in Act_S$ as a support node, where $h \in R^N$, and $q, s \in \{1, 2, ..., HW\}$. 
Then, we calculate the cosine similarity $e_{q,s}=cos(h_q, h_s)$ between all node pairs, with one from $Act_Q$ and the other from $Act_S$:
\begin{equation}\label{eq:eqs}
e_{q,s} = \frac{h_q^{T} h_s}{\left \| h_q \right \| \left \| h_s \right \|}\quad
q, s \in \{1, 2, ..., HW\}.
\end{equation}

For each query node $h_q$, we obtain an $H\times W$ similarity map $e_q$, which is then element-wise multiplied with the support mask to keep the similarity of the foreground support nodes while setting the similarity of the background support nodes to $-\infty$,
followed by Softmax to generate the weights: 
\begin{equation}
w_{q,s} = \frac{\exp(e_{q,s})}{\sum_{k=1}^{HW} \exp(e_{q, k})}.
\label{energy}
\end{equation}
We then do a weighted sum over all support node features and multiply it with the original query node feature: 
\begin{equation}
v_q = \sum_{s=1}^{HW} w_{q,s} h_{s},
\label{eq:vq}
\end{equation}
\begin{equation}
h_q' = h_q \odot v_q
\label{eq:hq}
\end{equation}

where $\odot$ denotes element-wise product. 
Combining all the fused query features $h_q'$, we obtain the fused activation map $Act_Q'\in \mathbb{R}^{H\times W\times N}$. Here, \eqref{eq:vq} essentially selects the most similar foreground support nodes and \eqref{eq:hq} highlights the query nodes matched with foreground support nodes while suppressing the query nodes matched with background support nodes, all in the context of the meta-class representation. 

\begin{figure*}[t]
\centering
    \includegraphics[width=0.7\linewidth]{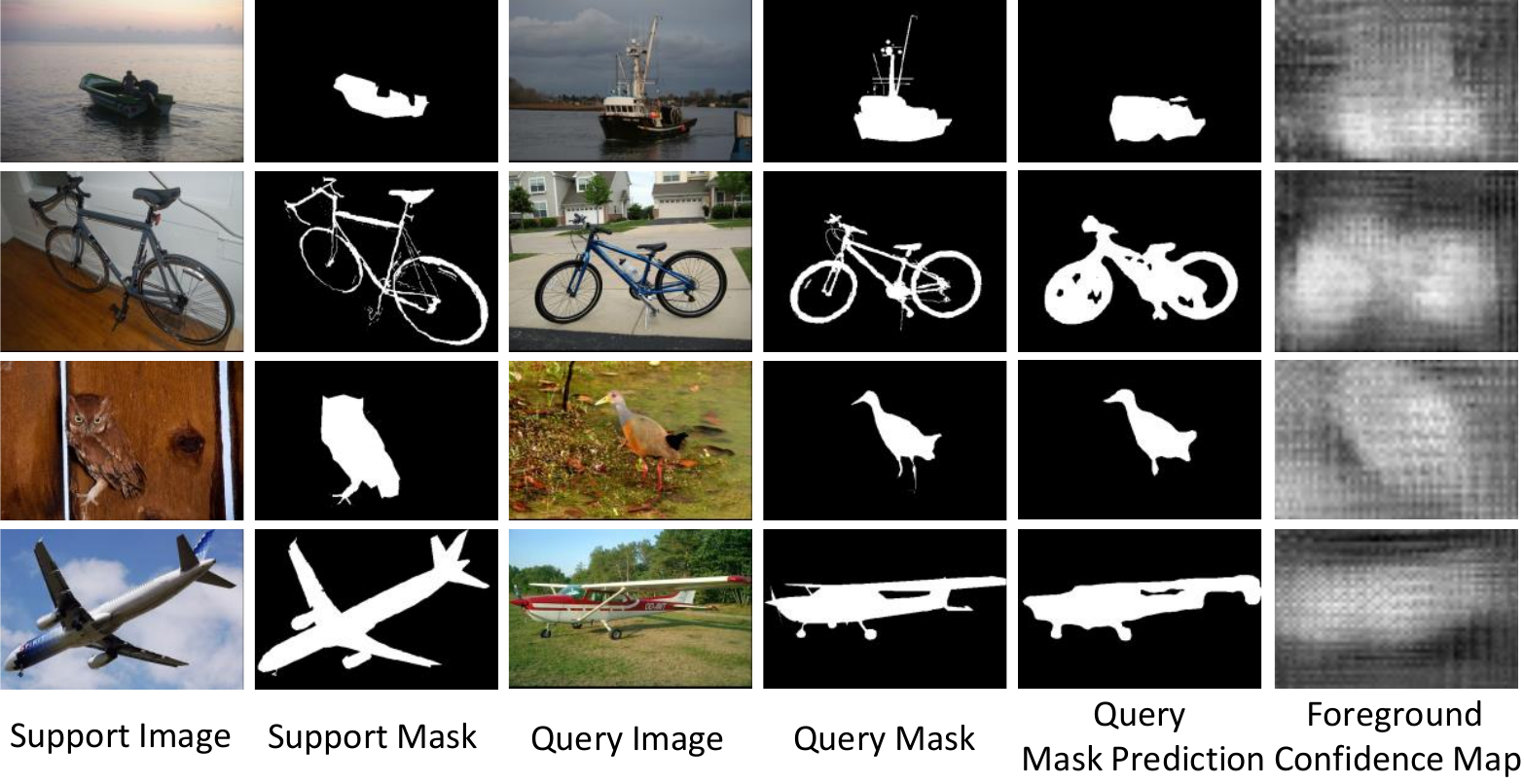}
    \caption{Visual results of our proposed MM-Net on fold-0 of PASCAL-$5^i$ dataset.}
    \label{results}
\end{figure*}

\subsection{Foreground Confidence Module}
Inspired by PFENet \cite{tian2020pfenet}, which concludes that the high level features can give a guidance mask telling the probability of pixels belonging to the target class. Thus, we further introduce a foreground confidence module (FCM) to produce a high-level foregournd confidence map. Compared with the previous process based on MMM and APM, which facilitates the interactions of the query and support images via middle-level meta-class features, FCM facilitates their interactions via high-level within-class features, i.e. the 4th level features of the pre-trained ResNet50. 

For simplicity, we reuse $F_Q$ and $F_S$ for the high level backbone feature maps for the query and support images, respectively. To generate the foreground confidence map $C_Q$, 
we first update $F_S$ by element-wise multiplying it with the support mask. Then, similar to that in APM, we compute cosine similarity $cos(f_q, f_s)$ between all pairs of feature nodes of $f_q \in F_Q$ and $f_s \in F_S$ as
\begin{equation}
cos(f_q, f_s) = \frac{f_q^{T} f_s}{\left \| f_q \right \| \left \| f_s \right \|}\quad
q, s \in \{1, 2, ..., HW\}.
\end{equation}
For each $f_q$, we take the maximum similarity among all support nodes as the foreground probability value $c_q \in \mathbb{R}$ as
\begin{equation}
c_q = \max\limits_{s\in \{1, 2, ..., HW\}}(cos(f_q, f_s)).
\end{equation}
We then reshape all the probability values $c_q$ into the foreground confidence map $C_Q  \in \mathbb{R}^{H\times W}$. Finally, 
we normalize all the values in $C_Q$ by a min-max normalization: \begin{equation}
C_Q = \frac{C_Q - \min(C_Q)}{\max(C_Q) - \min(C_Q) + \epsilon},
\end{equation}
where $\epsilon$ is set to $10^{-7}$. 

With the fused query attention map $Act_Q'$ from APM and the foreground confidence map $C_Q$ from FCM, we apply a channel-wise concatenation of the two maps, and then pass it to the feature enrichment module (FEM)~\cite{tian2020pfenet} to generate the final segmentation mask.

\subsection{Quality Measurement Module}
The diagram in Fig.~\ref{overall} is only for one-shot setting. When it comes to $K$-shot ($K>1$) settings, more than one support images are given. A common way is to average the features \cite{zhang2019pyramid} extracted from the support images and then pass the averaged feature for further processing. Such a simple average feature fusion might not be good, since some support images could be of poor quality for generating the class prototype features. Thus, we further propose a Quality Measurement Module (QMM) to select high quality support features. 

Specifically, we make use of the cosine similarity $e_{q,s}$ in~\eqref{eq:eqs} with $-\infty$ on the  background regions, same as in Section~\ref{Sec:APM}. For the $k$th support image, we have $e^k_{q,s}$. 
Then, we compute the quality measure for the $k$th support image and the $q$th query node:
\begin{equation} \label{eq:pkq}
p^k_q = \sum_{s=1}^{HW}(\sigma(e^{k}_{q,s})), \ k\in \{1, 2, ..., K\}, \  q \in \{1, 2, ..., HW\}
\end{equation}
where $\sigma(\cdot)$ is a Sigmoid function. Essentially, \eqref{eq:pkq} suggests that for the $q$th query node, the larger similarity sum from the the $k$th support image, the higher quality / weight we rank the support image. 
After that, we reshape $p^k_q$, $q \in \{1, 2, ..., HW\}$ into a map $P^k_Q  \in \mathbb{R}^{H\times W}$ aligning with the fused activation map $Act_Q'$. With the obtained $K$ maps $P^k_Q$, $k\in \{1, 2, ..., K\}$, we further apply softmax over the $k$ dimension to normalize the quality maps across different support images. Finally, we treat each quality map $P^k_Q$ as a weight map, multiply with the corresponding fused activation map $Act'^k_Q$, and sum them together. In this way, we obtain the final weighted average map $Act_Q'$, which is then passed to FEM for segmentation prediction.

\begin{table*}
\caption{1-shot and 5-shot mIoU results on PASCAL 5$^i$ dataset. We list the backbone and training size used by each method. 
Our MM-Net outperforms the state-of-the-art under all the experiment settings.}
\label{sota_voc}
\centering
\resizebox{0.95\linewidth}{!}{
\begin{tabular}{c|c|c|cccc|c|cccc|c}
\hline
\toprule[1.5pt]
{ }                          & { }                                & { }                           & \multicolumn{5}{c|}{{ 1-shot}}                                                                                                       & \multicolumn{5}{c}{{ 5 shot}}                                                                                                       \\ \cline{4-13} 
\multirow{-2}{*}{{ Methods}} & \multirow{-2}{*}{{ Training Size}} & \multirow{-2}{*}{{ Backbone}} & { Fold-0} & { Fold-1} & { Fold-2} & { Fold-3} & { Mean}          & { Fold-0} & { Fold-1} & { Fold-2} & { Fold-3} & { Mean}          \\ 
\midrule[1pt]
{ PANet \cite{wang2019panet}}                     & { 417 $\times$ 417 }                       & { VGG 16}                  & { 42.3}   & { 58.0}   & { 51.1}   & { 41.2}   & { 48.1}          & { 51.8}   & { 64.6}   & { 59.8}   & { 46.5}   & { 55.7}          \\ 

{ FWBF \cite{nguyen2019feature}}                     & { 512 $\times$ 512}                       & { VGG 16}                  & { 47.0}   & { 59.6}   & { 52.6}   & { 48.3}   & { 51.9}          & { 50.9}   & { 62.9}   & { 56.5}   & { 50.1}   & { 55.1}          \\ 
\midrule[0.5pt]

{ CANet \cite{zhang2019canet}}                     & { 321 $\times$ 321}                       & { ResNet 50}                  & { 52.5}   & { 65.9}   & { 51.3}   & { 51.9}   & { 55.4}          & { 55.5}   & { 67.8}   & { 51.9}   & { 53.2}   & { 57.1}          \\ 
{ PGNet\cite{zhang2019pyramid}}                     & { 321 $\times$ 321}                       & { ResNet 50}                  & { 56.0}   & { 66.9}   & { 50.6}   & { 50.4}   & { 56.0}          & { 54.9}   & { 67.4}   & { 51.8}   & { 53.0}   & { 56.8}          \\ 
{ CRNet \cite{liu2020crnet}}                     & { 321 $\times$ 321}                       & { ResNet 50}                  & { -}      & { -}      & { -}      & { -}      & { 55.7}          & { -}      & { -}      & { -}      & { -}      & { 58.8}          \\ 
{ PMMs \cite{yang2020prototype}}                       & { 321 $\times$ 321}                       & { ResNet 50}                  & { 55.2}   & { 66.9}   & { 52.6}   & { 50.7}   & { 56.3}          & { 56.3}   & { 67.3}   & { 54.5}   & { 51.0}   & { 57.3}          \\ 
{ PPNet \cite{liu2020part}}                     & { 417 $\times$ 417}                       & { ResNet 50}                  & { 47.8}   & { 58.8}   & { 53.8}   & { 45.6}   & { 51.5}          & { 58.4}   & { 67.8}   & \textbf{ 64.9}   & { 56.7}   & { 62.0}          \\ 
\midrule[0.5pt]
{ PFENet \cite{tian2020pfenet}}                    & { 473 X 473}                       & { ResNet 50 v2}               & { 61.7}   & { 69.5}   & { 55.4}   & { 56.3}   & { 60.8}          & \textbf{ 63.1}   & { 70.7}   & { 55.8}   & { 57.9}   & { 61.9}          \\ \midrule[0.5pt] \midrule[0.5pt]
{ Ours}                      & { 321 X 321}                       & { VGG}                        & { 57.1}   & { 67.2}   & { 56.6}   & { 52.3}   & { 58.3}          & { 56.6}   & { 66.7}   & { 53.6}   & { 56.5}   & { 58.3}          \\ 
{ Ours}                      & { 321 X 321}                       & { ResNet 50}                  & { 58.0}   & { 70.0}   & \textbf{ 58.0}  & { 55.0}   & { 60.2}          & { 60.0}   & { 70.6}   & { 56.3}   & { 60.3}   & { 61.8}          \\ 
{ Ours}                      & { 473 X 473}                       & { ResNet 50 v2}               & \textbf{ 62.7}   & \textbf{ 70.2}   & { 57.3}   & \textbf{ 57.0}   & { \textbf{61.8}} & { 62.2}  & \textbf{ 71.5}  & { 57.5}  & \textbf{62.4}  & { \textbf{63.4}} \\ \bottomrule[1.5pt]
\end{tabular}
}
\end{table*}

\subsection{Training Loss}
\textbf{Image Segmentation loss} is used to supervised the segmentation mask generation. Specifically, following PFENet~\cite{tian2020pfenet}, we apply multiple cross entropy losses, with $\mathcal{L}_{seg_2}$ on the final segmentation prediction $\hat{Y}_{Q}$ and $\mathcal{L}_{seg_1}^i$ ($i \in \{1, 2, ..., L\}$) on the intermediate masks $\hat{Y}_{Q}^{i}$. 

\textbf{Memory Reconstruction Loss.} To avoid the meta-class memory from learning similar embeddings, we propose a memory reconstruction loss function to encourage learning meaningful and diverse meta-class embeddings. Specifically, we firstly apply a channel-wise Softmax function over all the activation maps $Act^{n}(x,y)$ obtained in~\eqref{eq:act} as 
\begin{equation}\label{eq:hact}
    \hat{Act^{n}}(x,y) = \frac{\exp(Act^{n}(x,y))}{\sum_{k=1}^{N} \exp(Act^{k}(x,y)}).
\end{equation}
Then, we use  $\hat{Act^{n}}(x,y)$ and the meta-class embeddings $M$ to reconstruct the original image features $F(x,y)$ as 
\begin{equation} \label{eq:hf}
    \hat{F}(x,y) = \sum_{n=1}^{N} \hat{Act}^{n}(x,y)M^n .
\end{equation}
Essentially, \eqref{eq:hact} and \eqref{eq:hf} are to select the most similar meta-class embeddings to obtain the reconstructed feature $\hat{F}$. Reshaping $\hat{F}$ into $D\times HW$, we then compute the correlation matrix $C_f \in R^{HW\times HW}$:
\begin{equation}
    C_f = \hat{F}^T F.
\end{equation} 
Finally, we define the reconstruction loss $\mathcal{L}_{Recon}$ as a cross-entropy loss to
maximize the log-likelihood of the diagonal elements in $C_f$. This reconstruction loss encourages different meta-class embeddings to be different. This is because, if all $M^n$ are similar, it will not be able to well reconstruct the diverse original feature $F$. 

The overall loss function can be summarized as
\begin{equation}
\label{eqn:loss}
\mathcal{L} = \frac{\alpha}{L} \sum_{i=1}^L \mathcal{L}_{Seg_1} + \beta \mathcal{L}_{Seg_2}^i + \gamma\mathcal{L}_{Recon},
\end{equation}
where $\alpha$, $\beta$ and $\gamma$ are the trade-off parameters, being set as 1, 1 and 0.1, respectively.

\section{Experiments}

\subsection{Implementation details}
\noindent \textbf{Datasets.} We follow PFENet~\cite{tian2020pfenet} to conduct experiments on PASCAL-$5^i$~\cite{shaban2017one} and COCO~\cite{lin2014microsoft} datasets. PASCAL-$5^i$ combines PASCAL VOC 2012 with external annotations from SDS dataset~\cite{hariharan2014simultaneous}. It contains 20 classes, divided into 4 folds with 5 classes per fold. We randomly sample 5000 support-query pairs for testing. For COCO, following~\cite{tian2020pfenet}, we split its 80 classes into 4 folds with 20 classes per fold. The class indexes in fold $i$ are selected according to ${4x - 3 + i}$, where $x \in [1, 20]$ and $i \in [1, 4]$. We random select 20,000 support and query pairs for testing.

\noindent\textbf{Experiment setting.} For fair comparisons with previous methods, we consider multiple backbones including VGG-16, ResNet-50 and ResNet-50-v2. 
Here, VGG-16 and ResNet-50 are the commonly used backbone networks and ResNet-50-v2 is a modified version by PFENet~\cite{tian2020pfenet}, where the standard $7 \times 7$ convolution layers are replaced with a few $3 \times 3$ convolutional layers. All the backbone networks are pre-trained on the ImageNet classification task and fixed during our model training. We use SGD for training the rest of the network layers with momentum and weight decay being 0.9 and $10^{-4}$, respectively. In addition, we use a learning rate of 0.0025 and a batch size of 4 to train our model for both $1$-shot and $5$-shot settings. All of our experiments are conducted on one NVIDIA RTX 2080Ti GPU.

\noindent\textbf{Evaluation metrics.} Following the previous works~\cite{zhang2019canet,liu2020crnet}, we adopt the class mean intersection over union (mIoU) as our evaluation metric for ablation studies and final comparisons.

\begin{table*}
\caption{1-shot and 5-shot mIoU results on COCO dataset. The results of CANet* is obtained from~\cite{yang2020prototype}.}
\label{sota_coco}
\centering
\resizebox{0.95\linewidth}{!}{
\begin{tabular}{c|c|c|cccc|c|cccc|c}
\toprule[1.5pt]
{ }                          & { }                                & { }                           & \multicolumn{5}{c|}{{ 1-shot}}                                                                                                       & \multicolumn{5}{c}{{ 5 shot}}                                                                                                       \\ \cline{4-13} 
\multirow{-2}{*}{{ Methods}} & \multirow{-2}{*}{{ Training Size}} & \multirow{-2}{*}{{ Backbone}} & { Fold-0} & { Fold-1} & { Fold-2} & { Fold-3} & { Mean}          & { Fold-0} & { Fold-1} & { Fold-2} & { Fold-3} & { Mean}          \\  \midrule[0.5pt]
PANet \cite{wang2019panet}   &     417 $\times$ 417          & VGG 16     & -     & -                              & -    & -     & 20.9    & -    & -    & -    & -     & 29.7 \\ 
FWBF \cite{nguyen2019feature}     &     512 $\times$ 512  & ResNet 101 & 19.9  & 18.0                           & 21.0 & 28.9  & 21.2    & 19.1 & 21.5 & 23.9 & 30.1  & 23.7 \\ 
CANet* \cite{zhang2019canet}  & 321 $\times$ 321     & ResNet 50  & 25.1  & 30.3                           & 24.5 & 24.7  & 26.1    & 26.0 & 32.4 & 26.1 & 27.0 & 27.9 \\ 
PMMs \cite{yang2020prototype}   &   321 $\times$ 321 & ResNet 50  & 29.5  & 36.8 & 29.0 & 27.0  & 30.6    & 33.8 & 42.0 & 33.0 & 33.3  & 35.5 \\ 
PFENet \cite{tian2020pfenet}  & 641 $\times$ 641     & ResNet 101 v2 & 34.3  & 33.0 & 32.3 & 30.1  & 32.4    & 38.5 & 38.6 & 38.2  & 34.3  & 37.4 \\ \midrule[0.5pt] \midrule[0.5pt]
Ours    &  321  $\times$ 321       & ResNet 50  & 34.9  & 41.0                           & \textbf{37.8} & 35.2  & 37.2    & \textbf{38.5} & 39.6 & 38.4 & 35.5  & 38.0\\
Ours    &  473  $\times$ 473       & ResNet 50 v2  & \textbf{34.9}  & \textbf{41.0}                           & 37.2 & \textbf{37.0}  & \textbf{37.5}    & 37.0 & \textbf{40.3} & \textbf{39.3} & \textbf{36.0}  & \textbf{38.2} \\\bottomrule[1.5pt]
\end{tabular}
}
\end{table*}

\subsection{Comparisons with state-of-the-art}
Tables~\ref{sota_voc} and~\ref{sota_coco} show the 1-shot and 5-shot mIoU results of different methods on PASCAL-$5^i$ and COCO datasets, respectively. 
We list the training size and the backbone used in the previous methods. For CANet, PGNet, CRNet and PMM methods, they all use the image size of 321 $\times$ 321 with the standard ResNet-50 backbone to extract the features. However, PPNet and PFENet use larger image sizes. As observed in~\cite{chen2017rethinking}, a larger image size usually gives a better segmentation performance. Moreover, PFENet also uses a more powerful ResNet-50-v2 backbone. For a fair comparison, we report the performance of our method under different image sizes and backbones. We can see that our MM-Net achieves the best results under all training conditions.

For the experiments on COCO dataset, PFENet uses the image size of $641 \times 641$ (as specified in their released code) with ResNet-101-v2 as backbone. Due to our GPU memory restriction, we still use ResNet-50-v2 as our model backbone and $473 \times 473$ as our training size. Despite this, our 1-shot results still outperform PFENet by 5.1\%, as shown in Table~\ref{sota_coco}. In addition, Fig.~\ref{results} gives a few qualitative testing results on fold-0 of PASCAL-$5^i$ dataset.

For the inference speed and GPU memory consumption, our proposed MM-Net consumes 2466 MiB GPU memory (17 FPS), comparable to previous SOTA PFENet’s 1920 MiB GPU memory consumption (42 FPS). The slightly more memory consumption and running time of MM-Net are due to our introduced Meta-class Memory Module.

\subsection{Ablation studies}
\begin{figure*}[t]
\centering
    \includegraphics[width=0.6\linewidth]{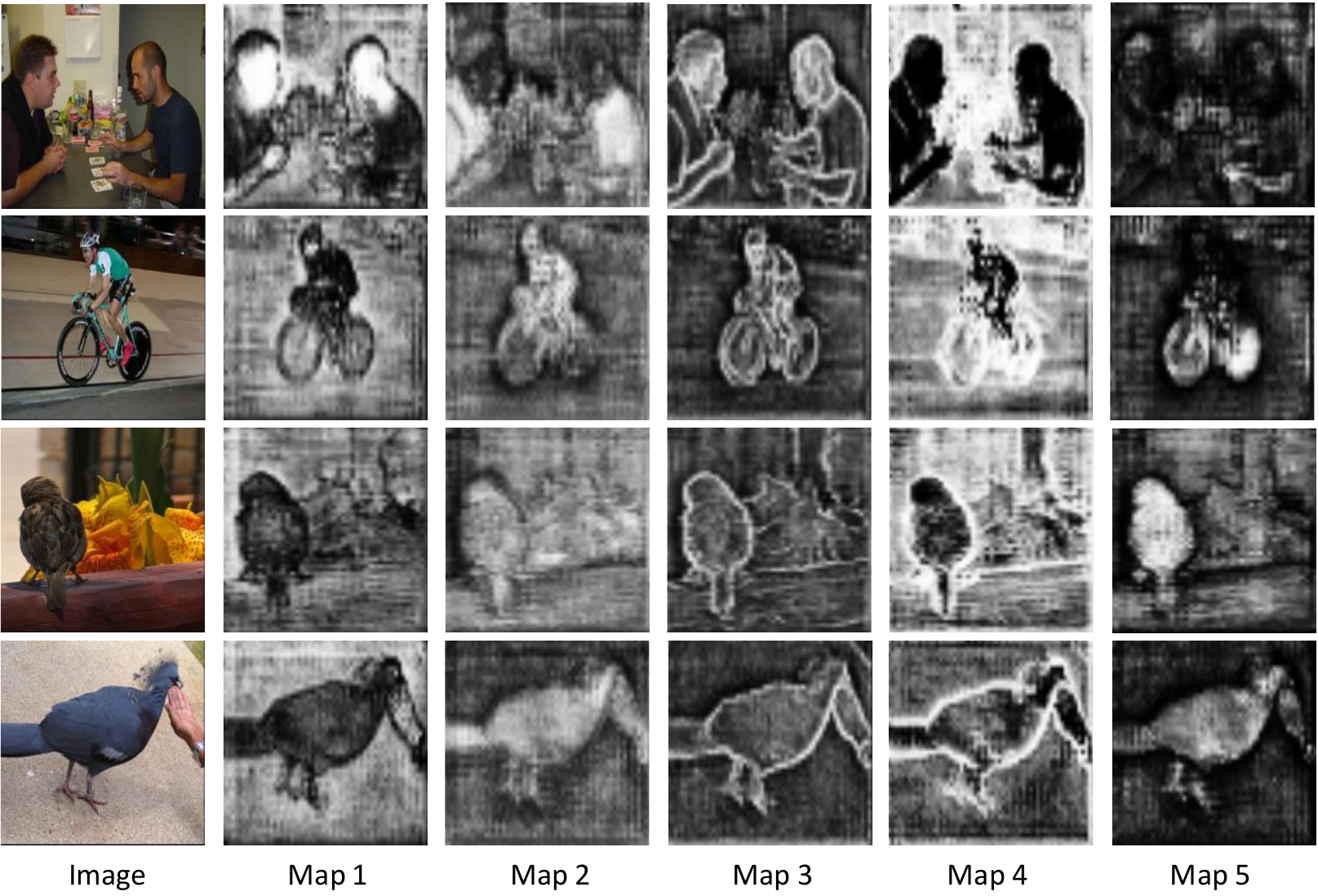}
    \caption{Visual results of the meta-class activation maps $Act$. We random select 5 from all 50 activation maps and all activation maps are obtained from the same meta-class memory. In the 1st row, we can see the meta-class memory highlights human's head, torso, edge, etc.}
    \label{AttnMap}
\end{figure*}

\begin{table}
\caption{Ablation studies on PASCAL-5$^i$ dataset about how many meta-class embeddings are better. }
\label{number_emb}
\resizebox{\linewidth}{!}{
\begin{tabular}{c|cccc|c}
\toprule[1.5pt]
\multicolumn{1}{l|}{Numbers of}          & \multicolumn{5}{c}{1 shot}              \\ \cline{2-6} 
\multicolumn{1}{l|}{Meta-class Embeddings} & fold 0 & fold 1 & fold 2 & fold 3 & Mean \\ \midrule[0.5pt]
20      & 61.4      & 70.0      & \textbf{57.5}      & \textbf{58.0}      & 61.7    \\
50      & \textbf{62.7}   & 70.2   & 57.3   & 57.0   & \textbf{61.8} \\
100     & 62.2      & \textbf{70.4}      & 55.9      & 55.6      & 61.0    \\ \bottomrule[1.5pt]
\end{tabular}
}
\end{table}

\begin{table}
\caption{Ablation studies of our proposed Meta-class Memory Module (MMM) on PASCAL 5$^i$ dataset.
 }
\label{TableMMM}
\resizebox{\linewidth}{!}{
\begin{tabular}{l|cccc|c}
\toprule[1.5pt]
\multirow{2}{*}{Methods}    & \multicolumn{5}{c}{1 shot}              \\  \cline{2-6} 
                & fold 0 & fold 1 & fold 2 & fold 3 & Mean \\ \midrule[0.5pt]
FCM                 & 40.4  & 47.4  & 42.9   & 40.5   & 42.8 \\  
FCM + Feat          & 59.6  & 68.0  & 54.5   & 53.5   & 58.9 \\
FCM + MMM           & \textbf{62.7}  & \textbf{70.2}  & \textbf{57.3}   & \textbf{57.0}   & \textbf{61.8} \\ \bottomrule[1.5pt]
\end{tabular}

}
\end{table}

\begin{table}
\caption{Ablation studies on PASCAL 5$^i$ dataset about which levels of features are better. Here (2+3) indicates the 2nd and 3rd level features are fused, and (2,3) indicates we learn two memories from the 2nd and 3rd level features,  respectively.}
\label{Tablelevel}
\resizebox{\linewidth}{!}{
\begin{tabular}{l|cccc|c}
\toprule[1.5pt]
\multirow{2}{*}{Methods}    & \multicolumn{5}{c}{1 shot}              \\  \cline{2-6} 
                & Fold 0 & Fold 1 & Fold 2 & Fold 3 & Mean \\ \midrule[0.5pt]

MMM(3) + APM        & 59.8   & 69.3   & 54.2   & 55.2   & 59.6 \\ 
MMM(2,3) + APM      & 60.2   & 70.0   & 53.7   & 56.0   & 60.0 \\ 
MMM(2+3) + Global   & 61.8   & 69.1   & 56.7   & 56.4   & 61.2 \\
MMM(2+3) + APM      & \textbf{62.7}   & \textbf{70.2}   & \textbf{57.3}  & \textbf{57.0}   & \textbf{61.8} \\ \bottomrule[1.5pt]
\end{tabular}

}
\end{table}

\begin{table}[]

\caption{Ablation studies on our proposed memory reconstruction loss under 1-shot setting on PASCAL 5$^i$ dataset. }
\label{memory_recon_loss}
\resizebox{\linewidth}{!}{
\begin{tabular}{cc|cccc|c}
\toprule[1.5pt]
\multicolumn{2}{l}{Recon Loss On:} & \multicolumn{5}{|c}{1 shot}              \\ \cline{3-7} 
Support           & Query          & Fold 0 & Fold 1 & Fold 2 & Fold 3 & Mean \\ \midrule[0.5pt]
                  &                & 60.7   & \textbf{70.2}   & 55.6   & 56.2   & 60.7 \\
                  & $\surd$              & \textbf{64.1}   & 69.8   & 56.1   & 55.5   & 61.4 \\
$\surd$                  & $\surd$               & 62.5   & \textbf{70.2}   & \textbf{57.7}   & 56.1   & 61.6 \\
$\surd$                  &                & 62.7   & \textbf{70.2}   & 57.3   & \textbf{57.0}   & \textbf{61.8} \\ \bottomrule[1.5pt]
\end{tabular}
}
\end{table}

\begin{table}
\caption{Ablation studies on our proposed Quality Measurement Module under 5-shot setting on PASCAL 5$^i$ dataset.}
\label{TableQMM}
\resizebox{\linewidth}{!}{
\begin{tabular}{l|cccc|c}
\toprule[1.5pt]
\multirow{2}{*}{Methods}    & \multicolumn{5}{c}{5 shot}              \\  \cline{2-6} 
                & Fold 0 & Fold 1 & Fold 2 & Fold 3 & Mean \\ \midrule[0.5pt]
Ours $w/o$ QMM                    & 60.7  & 71.0   & 56.9   & 61.8   & 62.6 \\
Ours $w/$ QMM              & \textbf{62.2}   & \textbf{71.5}   & \textbf{57.5}   & \textbf{62.4}   & \textbf{63.4} \\ 
\bottomrule[1.5pt]
\end{tabular}
}
\end{table}

\noindent\textbf{Number of Mate-class Memory Embeddings.} We conduct ablation experiments to analyze how many meta-class memory embeddings are better for memory learning. Table~\ref{number_emb} shows that 50 embeddings yield the best performance. This suggests that the network learns more meaningful and effective features with 50 meta-classes. Thus, we use 50 meta-class memory embeddings for our following experiments.

\noindent\textbf{Effect of Meta-class Memory Module.}
We construct two baselines to show the effectiveness of our proposed MMM. The first baseline is that we only use the foreground confidence map for the mask decoding (denoted as `FCM' in Table~\ref{TableMMM}). 
The 2nd baseline is that we add the middle-level image features for the mask prediction. Specifically, we extract the middle-level features (2nd and 3rd levels) of support and query images. Instead of computing the meta-class activate maps, we direct pass the image features to APM and FEM to predict the query mask. This baseline is denoted as `FCM+Feat' in Table~\ref{TableMMM}. 

As can be seen in Table~\ref{TableMMM}, our proposed method `FCM+MMM' improves the segmentation performance with 2.9\% gain in mean mIoU compared with directly using the middle-level features for decoding (`FCM+Feat'). This suggests that our proposed meta-class memory module is able to give better class prototype information for the query mask prediction.

Fig.~\ref{AttnMap} gives some examples of the computed meta-class activation maps $Act$, where we random select 5 from all 50 meta-class activation maps and all the maps are obtained from the same learned meta-class memory. As we can see, different meta-class embeddings memorize different meta-class features and capture different patterns in the images, e.g., capturing human head, torso, edge, etc, in the first row.

\noindent\textbf{Features for memory learning.} We conduct ablation experiments to analyze which level of feature is better for the memory learning. Table~\ref{Tablelevel} shows that fusing 2nd and 3rd level features (`2+3') yields the best performance. Our conjecture is that because the 2nd-level features capture more edge information and the 3rd-level features capture more part and object information, a combination of them leads to learning better meta-class memory embeddings. 

\noindent\textbf{Effect of Activation Propagation Module.} With the meta-class activation maps of query and support images, instead of using APM, one simple way to propagate the support mask information to query activation maps is to apply a global average pooling within the foreground region on the support activation maps to get a global average foreground representation vector, and then element-wise multiply it with each node in the query activation maps. This is denoted as `MMM(2+3)+Global' in Table~\ref{Tablelevel}. It can be seen that the APM is able to improve the mIoU by 0.6\%.

\noindent\textbf{Effect of Memory Reconstruction Loss.} Table~\ref{memory_recon_loss} shows the ablation study on our proposed memory reconstruction loss. It can be seen that the results with the loss are much better than the one without using the loss, clearly demonstrating its effectiveness. Note that the reconstruction loss can be applied for the reconstructions of different features, including query features, support features and both features. We can see that applying the loss for support features leads to the overall best performance. 

\noindent\textbf{Effect of Quality Measurement Module.}
Table~\ref{TableQMM} shows the effectiveness of our proposed Quality Measurement Module. For the baseline method (ours $w/o$ QMM), we use APM to obtain 5 fused activation maps independently from 5 different support images, and then we follow the conventional way to  average the 5 maps and pass the averaged map to FEM for the segmentation mask generation. Our proposed QMM improves mIoU by 0.8\%. This suggests that high quality support samples are more helpful for the query segmentation mask prediction.

\section{Conclusion}
In this paper, we have proposed a novel Meta-class Memory based few shot semantic segmentation method (MM-Net) with the major components of MMM, APM, FCM and QMM. The key novelty of our method lies in MMM, where we introduced a set of learnable meta-class embeddings to allow the common knowledge transfer between base classes and novel classes. Another novelty is from QMM,  
which can measure the quality of each support image so as to better fuse the support features. With all these components, our MM-Net has significantly improved the SOTA results on both PASCAL-5$^i$ and COCO datasets.

{\small
\bibliographystyle{ieee_fullname}
\bibliography{egbib}
}

\appendix\onecolumn

\renewcommand{\thetable}{\Alph{table}}
\setcounter{table}{0}
\renewcommand{\thefigure}{\Alph{figure}}
\setcounter{figure}{0}

\begin{center}
\textbf{\Large Learning Meta-class Memory for Few-Shot Semantic Segmentation} \\[5pt]

\textbf{\Large Supplementary Material} \\

\end{center}

\section{Effect of Foreground Confidence Module.}
Table~\ref{foreground} shows the effectiveness of the Foreground Confidence Module (FCM). The baseline method is that we directly pass the fused meta-class activate maps to FEM for the query mask prediction (denoted as `Ours $w/o$ FCM'). As shown, the FCM improves mIoU by 2.9\%, which suggests that high level features are helpful for the query segmentation mask prediction.

\begin{table}[h]
\centering
\caption{Ablation studies on the Foreground Confidence Module (FCM) under 1-shot setting on PASCAL 5$^i$ dataset.}
\label{foreground}
\resizebox{0.5\linewidth}{!}{
\begin{tabular}{l|cccc|c}
\toprule[1.5pt]
\multirow{2}{*}{Methods}    & \multicolumn{5}{c}{1 shot}              \\  \cline{2-6} 
                & Fold 0 & Fold 1 & Fold 2 & Fold 3 & Mean \\ \midrule[0.5pt]
Ours $w/o$ FCM                    & 57.5  & 68.7   & 55.8   & 53.7   & 58.9 \\
Ours $w/$ FCM              & \textbf{ 62.7}   & \textbf{ 70.2}   & \textbf{ 57.3}   & \textbf{ 57.0}   & { \textbf{61.8}} \\ 
\bottomrule[1.5pt]
\end{tabular}
}
\end{table}

\section{Comparison between Quality Measurement Module and the attention mechanism.} 

We further implement the attention mechanism of CANet into our method. Specifically, same as CANet, we concatenate the support and query features and pass them to two convolutional layers and a global average pooling layer to obtain the importance weights. The weights are used to fuse class activation maps and generate the final weighted activation map for final mask prediction. As shown in Table~\ref{attention}, with the attention mechanism of CANet (denote as ``Ours $w/$ Attn''), our model obtains mIoU of 61.7\% in the PASCAL $5^i$ dataset, which is lower than ours with QMM (63.4\%). This indicates that QMM is more suitable for our method.

\begin{table}[h]
\centering
\caption{Comparsion between our QMM and attention mechanism under 5-shot setting on PASCAL 5$^i$ dataset.}
\label{attention}
\resizebox{0.5\linewidth}{!}{
\begin{tabular}{l|cccc|c}
\toprule[1.5pt]
\multirow{2}{*}{Methods}    & \multicolumn{5}{c}{5 shot}              \\  \cline{2-6} 
                & Fold 0 & Fold 1 & Fold 2 & Fold 3 & Mean \\ \midrule[0.5pt]
Ours $w/$ Attn                    & 60.1  & 69.8   & 55.9   & 61.1   & 61.7 \\
Ours $w/$ QMM              & \textbf{62.2}   & \textbf{71.5}   & \textbf{57.5}   & \textbf{62.4}   & \textbf{63.4} \\ 
\bottomrule[1.5pt]
\end{tabular}
}
\end{table}

\section{Visualization results for ablation studies ``Features for memory learning''}

\begin{figure}[ht]
\centering
    \includegraphics[width=0.8\linewidth]{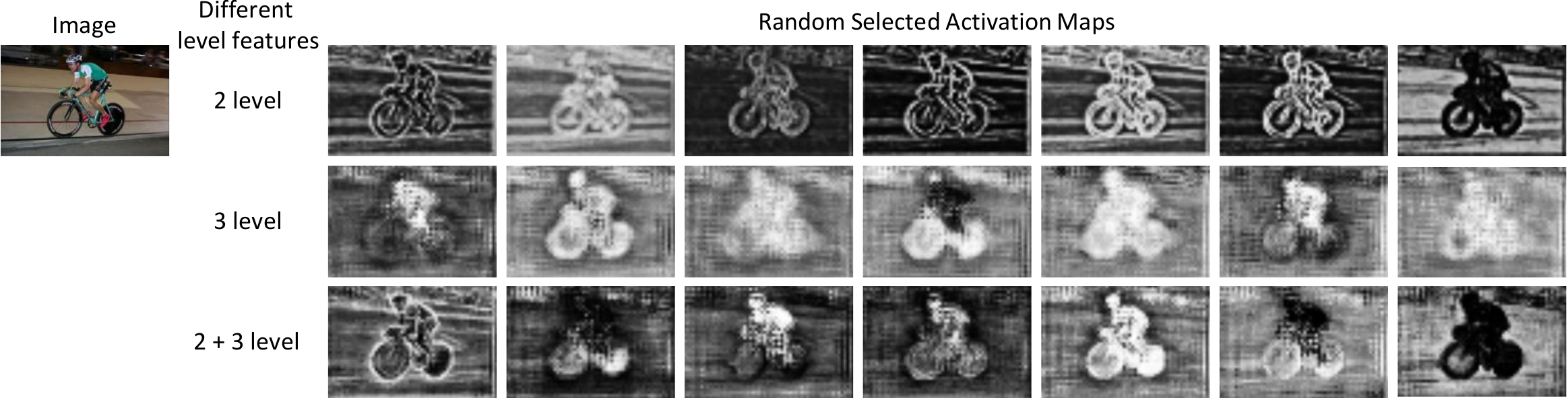}
    \caption{Visual results of the meta-class activation maps with different levels of image features. Here, $2 + 3$ refers to the fused features by channel-wise concatenating the 2nd and 3rd level features followed by a convolution. A combination of them leads the memory to have high activation on the meta-class regions (e.g. edge, part, object, etc.)}
    \label{multi-level-attn}
\end{figure}

We visualize the meta-class activate maps obtained from different level features in Fig~\ref{multi-level-attn}. As shown in the figure, the memory learned from 2nd level features has high activation on local information (e.g. edge) and the one from 3rd level features has high activation on the whole object or parts. A combination of them leads to learning better meta-class memory embeddings, which are able to have high activation on meta-class regions.

\end{document}